\documentclass[a4paper]{article}
\usepackage{INTERSPEECH2018}

\usepackage{multicol}
\usepackage{graphicx}
\usepackage{multirow}
\usepackage{amssymb,amsmath,bm}
\usepackage{enumitem}
\usepackage{color}
\usepackage{array}
\usepackage{footnote}
\makesavenoteenv{table}
\makesavenoteenv{algorithm}
\usepackage{algorithm}
\usepackage[noend]{algpseudocode}
\usepackage{hyperref}
\usepackage{color}

\usepackage{listings}             

\title{Linguistic Search Optimization for Deep Learning Based LVCSR}
\name{Zhehuai Chen
}
\address{
  SpeechLab, Department of Computer Science and Engineering, Shanghai Jiao Tong University}
\email{chenzhehuai@sjtu.edu.cn}

\begin{document}


\definecolor{mygreen}{rgb}{0,0.5,0}
\definecolor{mygray}{rgb}{0.5,0.5,0.5}
\definecolor{mymauve}{rgb}{0.58,0,0.82}
\lstset{ %
backgroundcolor=\color{white},   
basicstyle=\footnotesize\ttfamily,        
columns=fullflexible,
breaklines=true,                 
captionpos=b,                    
tabsize=4,
commentstyle=\color{mygreen},    
escapeinside={\%*}{*)},          
keywordstyle=\color{blue},       
stringstyle=\color{mymauve}\ttfamily,     
frame=single,
language=c++,
}

\maketitle
%

\vspace{-0.40em}
\section{Motivation and General Ideas}
\label{sec:intro}
\vspace{-0.40em}

Recent advances in deep learning based {\em large vocabulary continuous speech recognition} (LVCSR)  invoke
growing demands in large scale speech transcription. 
The inference process of a speech recognizer is to find a sequence of labels whose corresponding acoustic and language models best match the input feature~\cite{Huang:2001:SLP:560905}. The main computation includes two stages: acoustic model (AM) inference and linguistic search (weighted finite-state transducer, WFST). Large computational overheads of both stages hamper the wide application of LVCSR.

To reduce computation of the first stage, researchers have proposed a variety of efficient forms of AMs,
including novel structures~\cite{xue2014singular,peddinti2018low}, quantization
~\cite{mcgraw2016personalized,xiang2017binary} and frame-skipping~\cite{pundak2016lower}.
Meanwhile, algorithmic improvement of linguistic search, e.g. pruning~\cite{mohri2002weighted}, rescoring~\cite{hori2004fast}
and lookahead~\cite{soltau2009dynamic,nolden2012search}, was the mainstream approach to speed up the second stage in the past. 
%

This work mainly focuses on  linguistic search. 
Despite the WFST based LVCSR approach has been improved for several decades, two fundamental deficiencies remain:
i) The WFST search space is large and its graph traversal algorithms are conducted at each frame, e.g. {\em frame synchronous Viterbi beam decoding} (FSD).
ii) Most of these algorithms are originally serial algorithms and parallelizing them is non-trivial.

Benefit from stronger classifiers,  deep learning, and more powerful computing devices, we propose general ideas and some initial trials to solve these fundamental problems:
\begin{itemize}[leftmargin=*]
	\item {\bf Reduce the search complexity by end-to-end modeling}
\end{itemize}
Recent advances in more potent neural networks enable stronger modeling effects in the context and the history of the sequential modeling~\cite{sak2014long,qian2016very,chen2018progressive,chen2018sequence,huang2018ctc}.
More labeled data further alleviates the sparseness and generalization problem in the modeling.
Thus, it is promising to decompose the sequence into larger model granularities.
Research has been conducted on different model granularities from frame level to the whole sequence~\cite{amodei2015deep,soltau2016neural,collobert2016wav2letter,sak2015fast,chan2016end}.
e.g., in~\cite{soltau2016neural}, a word level deep learning based acoustic model is trained on 125K hours labeled data and outperforms models with smaller granularity.
We propose to change the frequency of Viterbi search from each feature frame  to each label output.
Correspondingly, a post-processing is applied on the frame level acoustic model outputs to obtain the label outputs: i) Decide whether there is a label output at the current frame or not. ii) If so, conduct the search process. If not, discard the current output. Thus the post-processing can be viewed as  the approximated probability calculation of each output label. Based on this framework, the larger model granularities we take, the less search complexity we obtain.
\begin{itemize}[leftmargin=*]
\item {\bf Accelerate the search speed using parallel computing}
\end{itemize}
GPU-based parallel computing is another potential direction which utilizes a large number of  units to parallelize the computation.
As common language models (LM) can be expressed as WFSTs, the idea is to parallelize WFST graph traversal algorithms.
Our initial work parallelizes Viterbi algorithm~\cite{forney1973viterbi} and redesigns it to fully utilize parallel computing devices nowadays, e.g. Graphics Processing Units (GPU) and Field-Programmable Gate Arrays (FPGA). 
To utilize large LMs in the 2nd pass and support rich post-processing, our design is to decode the WFSTs and generate exact lattices~\cite{povey2012generating}. The decoder remains to be general-purpose and does not impose special requirements on the form of AM or LM. 
The ideas to apply GPU parallel computing in WFST decoding include:
i)  Abstract the dynamic programming in the Viterbi algorithm as thread synchronization using atomic GPU operations. ii)  Propose a load balancing strategy for parallel WFST search scheduling among GPU threads. iii) Parallelize exact lattice generation and pruning algorithms.
Similar ideas can be applied to other WFST algorithms, e.g. determinization and minimization~\cite{mohri2002weighted}, and speedups of them can be expected. 

\vspace{-0.40em}
\section{Label Synchronous Framework}
\label{sec:lsd}
\vspace{-0.40em}


The search process is proposed to change from the feature level  to the label level, i.e. {\em{label synchronous decoding}} (LSD)~\cite{zhc00-chen-is16,zhc00-chen-tasl2017}.
Within the label inference,  a post-processing is applied on the frame level  acoustic model outputs. The formulation and implementation are discussed on CTC~\cite{graves2006connectionist} and LF-MMI~\cite{povey2016purely}.

During inference stage, Viterbi beam search of CTC model~\cite{graves2006connectionist} can be expressed as,
\begin{equation} \label{eq:ctc-dec-lsd}
\vspace{-0.5em}
   \mathbf{w}^* = \mathop{\arg\!\max}\limits_\mathbf{w} \left\{
        P(\mathbf{w})
        \mathop{\max}\limits_{\mathbf{l}_\mathbf{w}} \frac{ \prod_{l\in\mathbf{l}_\mathbf{w}} P(l|\mathbf{x}) }{P(\mathbf{l}_\mathbf{w})}\right\}
     \end{equation}
     where $\mathbf{x}$ is the feature sequence, $\mathbf{w}$ is a word sequence and
${\mathbf{w}}^*$ is the best word sequence.
$\mathbf{l}_{\mathbf{w}}$ denotes the label sequence, e.g. the phoneme sequence, corresponding to $\mathbf{w}$. 
Within the calculation of $P(l|\mathbf{x})$, a post-processing is proposed on the frame level neural network outputs, $P(\pi|\mathbf{x})$.
Here, the set of {\em common $\tt blank$} frames are defined as: 
	$
    U=\{u:y^{u}_{\tt{blank}} > \mathcal{T}\}
    $, where $y^{u}_{\tt{blank}}$ is the probability of the $\tt blank$ unit at frame $u$. 
    With a softmax layer in the CTC model, if the $\tt blank$ acoustic score is large enough and approaching a constant of $1$, it can be regarded that all competing paths share the same span of the $\tt blank$ frame. Thus ignoring the scores of the  frame does not affect the acoustic score rank in decoding:
       \begin{equation} \label{eq:viterbi-blk-ctc}
  \begin{split}
      P(l|\mathbf{x})
      \triangleq\sum_{\pi\in\mathcal{B}(\mathbf{l})}
          \ \prod_{\pi}P(\pi|\mathbf{x})\\
        \simeq\sum_{\pi\in\mathcal{B}(\mathbf{l})}
         \  \prod_{\pi\in U}{y_{b_l}^u}{\prod_{\pi\not\in U}{y_{p_l}^u}}
         \simeq\sum_{\pi\in\mathcal{B}(\mathbf{l})}
         \  {\prod_{\pi\not\in U}{y_{p_l}^u}}
        \end{split}
       \end{equation}
where $\mathcal{B}$ is a one-to-many mapping between labels, e.g. phonemes, and CTC states. LSD is summarized as Algorithm~\ref{code:lsd-dsm-alg}. The main difference compared with  FSD Viterbi algorithm is the introduction of $isBlankFrame(F)$ to detect whether a frame is $\tt blank$ or not.
Recently, novel HMM topology was proposed in~\cite{povey2016purely,pundak2016lower}, which holds a similar one-to-many mapping as  $\mathcal{B}$ function of CTC. The three-state HMM contains a state simulating the function of $\tt blank$. Thus similar post-processing as Equation~(\ref{eq:viterbi-blk-ctc}) and its corresponding algorithm can be derived.

\vspace{-0.5em}
\begin{algorithm}[ht]
\vspace{0mm}
\caption{Label Synchronous Viterbi Beam Search for CTC \textcolor[rgb]{0,0.5,0}{(Inputs: start and end nodes, token queue, time frames)}}
\label{code:lsd-dsm-alg}
\begin{algorithmic}[1]
\Procedure{ LSD for CTC } {S, E, Q, T}
\State $Q \leftarrow S$ \Comment \textcolor[rgb]{0,0.5,0}{initialization with start node}
\For {each $t\in [1,T]$}    \Comment \textcolor[rgb]{0,0.5,0}{frame-wise NN Propagation}
\State $F \leftarrow NNPropagate(t)$
\If {\textcolor[rgb]{0.8,0,0}{!isBlankFrame($F$)}}   \Comment \textcolor[rgb]{0,0.5,0}{phone-wise decoding}
\State  $Q\leftarrow ViterbiBeamSearch(F, Q)$
\EndIf
\EndFor
\State $\hat B\leftarrow finalTransition(E,S,Q)$ \Comment \textcolor[rgb]{0,0.5,0}{to reach end node}
\State backtrace($\hat B$)
\EndProcedure
\end{algorithmic}
\end{algorithm}
\vspace{-0.5em}

\begin{figure}[tbhp!]
        \centering
        \vspace{-1em}
        \includegraphics[width=\linewidth]{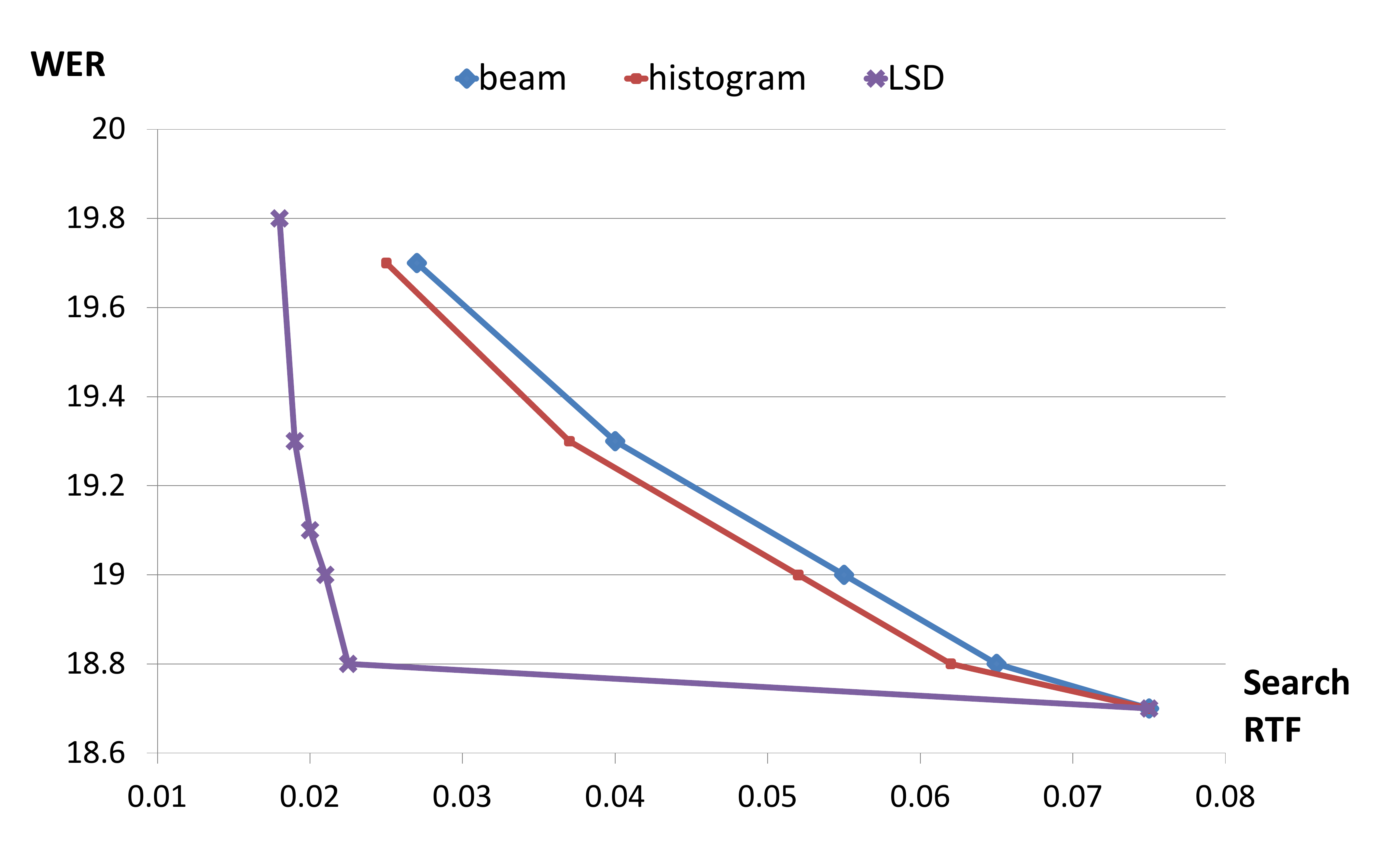}
        \vspace{-2em}
        \caption{{\it hub5e-swb WER versus Search Real-time Factor (RTF) in CTC Obtained by LSD and Other Pruning Methods. }} 
        \label{fig:prune-wer-at-dsm}
        \vspace{-1.5em}
      \end{figure}

The decoding complexity reduction from FSD to LSD is as Equation~(\ref{equ:complex-lsd}), where $T$ is the number of frames, $|L'|$ and $|W|$ are  sizes of label set and vocabulary. The number of $\tt blank$ frames, $|U|$, is always approaching T. Thus FSD is greatly sped up. 
\begin{equation}
\vspace{-0.5em}
\label{equ:complex-lsd}
\begin{split}
\mathbb{C} \propto\ T\cdot|L'| \cdot|W|\ \ \ \Rightarrow\ \ \ (T-|U|)\cdot|L'| \cdot|W|
\end{split}
\end{equation}
Experiments on Switchboard~\cite{godfrey1992switchboard} show the speedup as Figure~\ref{fig:prune-wer-at-dsm}.

\vspace{-0.40em}
\section{GPU-based Parallel WFST Decoding}
\label{sec:gpu-para}
\vspace{-0.40em}

\begin{figure}[ht]
  \centering
  \vspace{-1.7em}
    \includegraphics[width=1.1\linewidth]{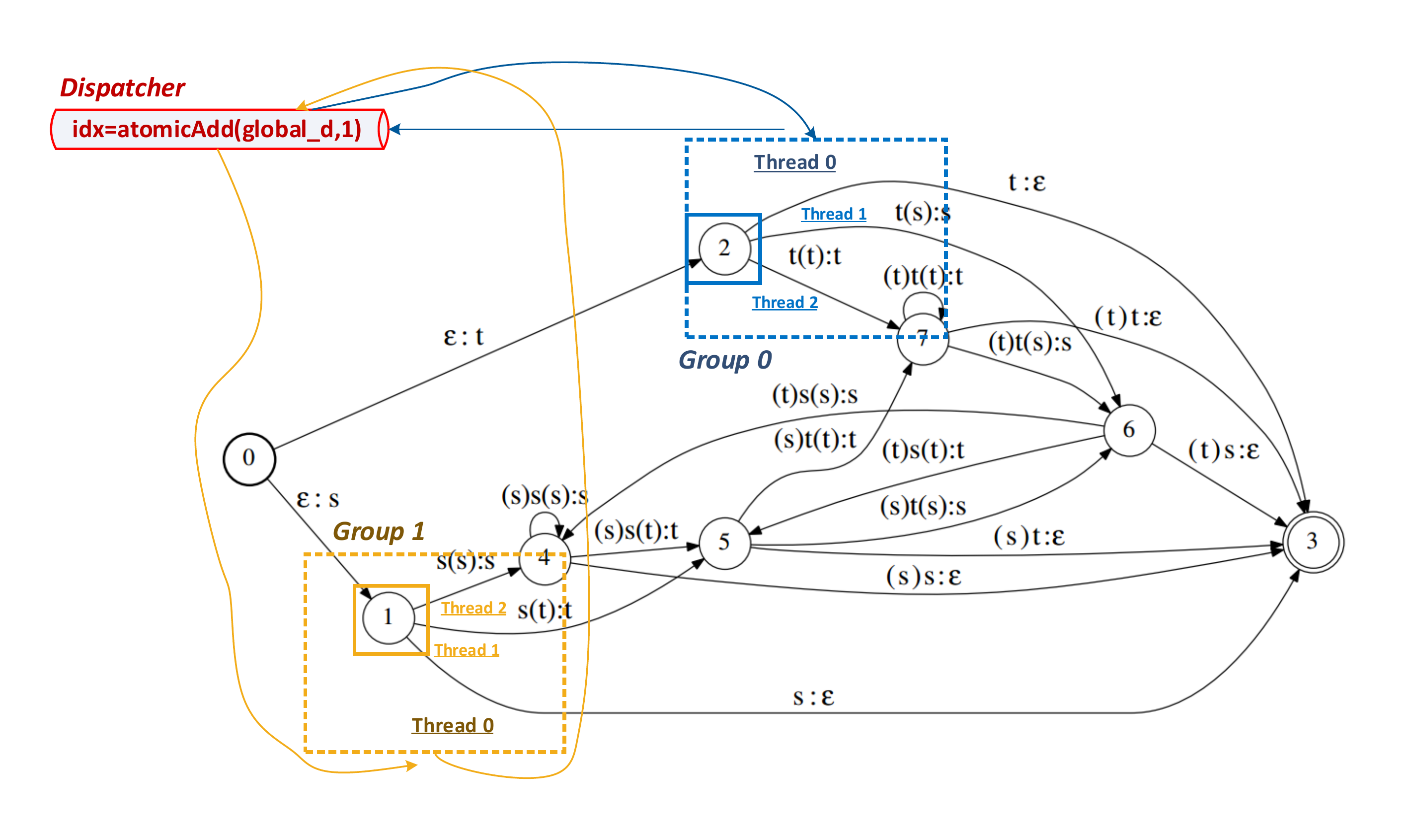}
    \vspace{-2em}
    \caption{\it Example of Dynamic Load Balancing. The dashed box denotes a
      CUDA cooperative group and different groups are with different colors.
        Each group is controlled by Thread 0 of it. After processing all the
        forward links of a state, Thread 0 accesses the dispatcher and the next token  is dynamically decided by atomic operation. Group 0 and 1 work in parallel. }
    \vspace{-1em}
    \label{fig:load-balance}
\end{figure}

Figure~\ref{fig:load-balance} shows the framework of parallel Viterbi beam search~\cite{chen2018gpu}.
%
The procedure of decoding is similar to the CPU version~\cite{povey2011kaldi}
but works in parallel with specific designs. 
Load balancing controls the thread parallelism over both WFST states and arcs.
Two GPU concurrent streams perform decoding and lattice-pruning in parallel 
launched by CPU asynchronous calls.

We parallelize {\em token passing algorithm}~\cite{woodland19951994} in two levels.
Tokens in different states are processed parallelly.
For each token, we traverse its out-going arcs in parallel as well.
Because WFST states might have different numbers of out-going arcs, the allocation
of states and arcs to threads can result in load imbalance.
We use a dispatcher in charge of global scheduling, and make $N$
threads as a group ($N = 32$) to process  arcs from a
token. When the token is processed, the group requests a new token from the dispatcher.
We implement task dispatching as an atomic operation~\cite{cuda9}. Figure~\ref{fig:load-balance}
shows an example.

At each frame, the Viterbi path is obtained  by 
a \textit{token recombination} procedure, where
a \textit{min} operation is performed on each state over all of its incoming arcs (e.g. state 7 in Figure~\ref{fig:load-balance} and the incoming arcs from state 2, 5 and 7), to compute the best cost and the corresponding predecessor of that state. 
We abstract this process as thread synchronization using atomic GPU operations. After finishing all synchronization, we aggregate survived tokens exploiting thread parallelism. We also parallelize exact lattice generation and pruning algorithms with similar ideas, described in~\cite{chen2018gpu}.

Experiments on Switchboard show that the proposed GPU decoder significantly and consistently speeds up CPU counterparts in varieties of GPU architectures, LMs and AMs. The implementation of this work is open-sourced~\footnote{\url{https://github.com/chenzhehuai/kaldi/tree/gpu-decoder}}.

\begin{figure}[h]
  \centering
  \vspace{-3.5em}
    \includegraphics[width=\linewidth]{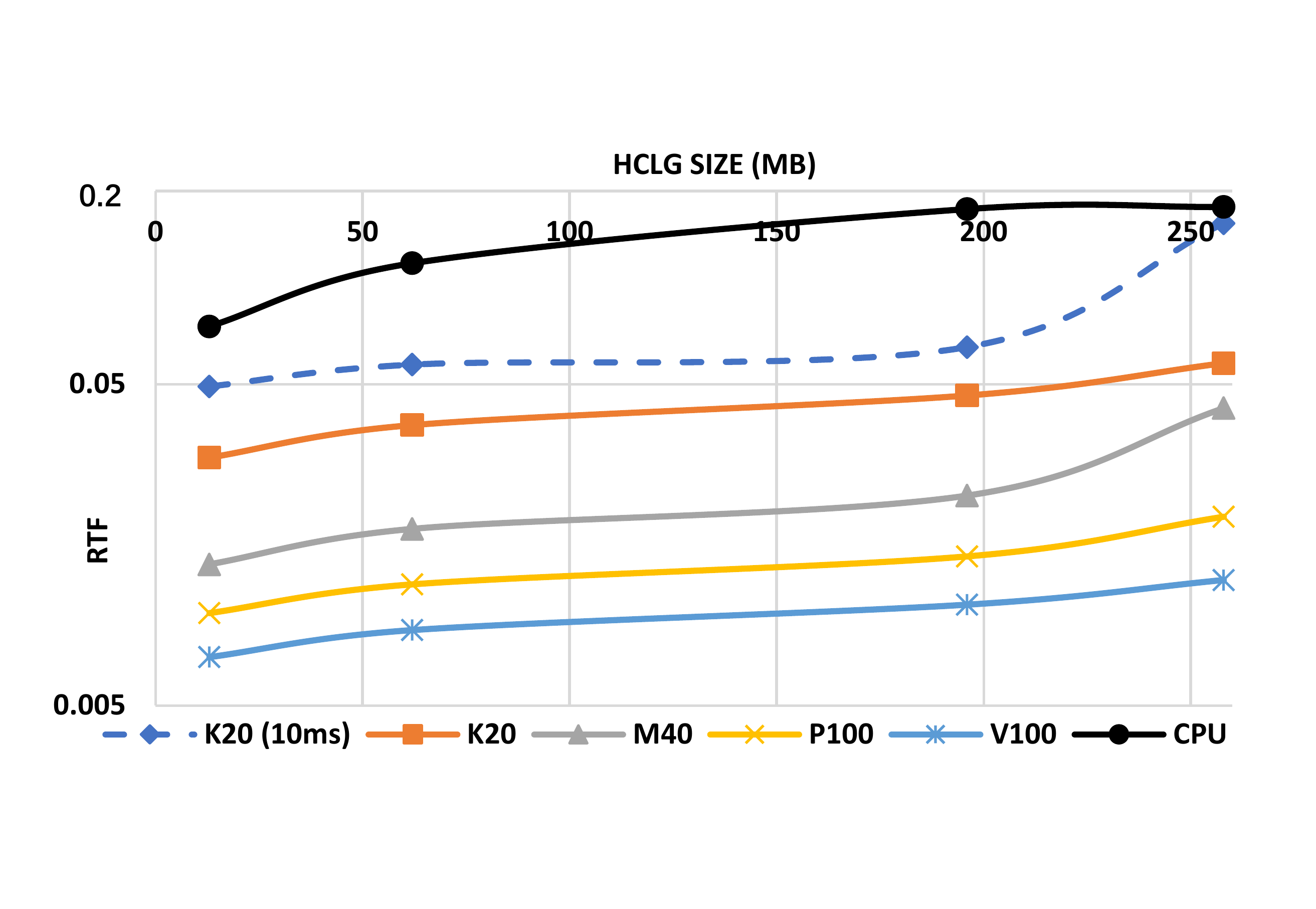}
    \vspace{-4.5em}
    \caption{\it  LM Size, Frame Rates and Architectures Comparison.}
    \vspace{-1.5em}
    \label{fig:exp-ana}
\end{figure}

\vspace{-0.40em}
\section{Future works}
\label{sec:future}
\vspace{-0.40em}

Our general ideas to reduce the computation of speech applications are two folds: reduce the search complexity by end-to-end modeling and accelerate WFST algorithms using parallel computing. For the first part, LSD in both discriminative and generative sequence models should be fully investigated. Integrating LSD as a sub-sampling module in the sequence-to-sequence framework is another direction~\cite{e2e-2018}. Keyword spotting and confidence measures in this framework needs to be considered~\cite{chen2017unified,chen2017confidence,chen2018kws}. For the second part, the initial work in Viterbi decoding can be extended to most of WFST graph traversal algorithms~\cite{mohri2002weighted}. According to specific application scenarios, parallel implementations of various devices (e.g. CPU, GPU and FPGA) can be considered. Neural network language models should be taken into account~\cite{xu2017pruned}.

\vspace{-0.40em}
\section{Acknowledgements}
\label{sec:ack}
\vspace{-0.40em}

I express sincere gratitude to my advisors, Kai Yu and Yanmin Qian. 
I thank the speech technology group of AISpeech Ltd.
for infrastructure support and valuable technical discussions.
I thank my internship mentors, Jasha Droppo, Jinyu Li, Daniel Povey, Justin Luitjens, Christian Fuegen and Yongqiang Wang.

\bibliographystyle{IEEEtran}

\bibliography{mybib}

\begin{thebibliography}{10}
\providecommand{\url}[1]{#1}
\csname url@samestyle\endcsname
\providecommand{\newblock}{\relax}
\providecommand{\bibinfo}[2]{#2}
\providecommand{\BIBentrySTDinterwordspacing}{\spaceskip=0pt\relax}
\providecommand{\BIBentryALTinterwordstretchfactor}{4}
\providecommand{\BIBentryALTinterwordspacing}{\spaceskip=\fontdimen2\font plus
\BIBentryALTinterwordstretchfactor\fontdimen3\font minus
  \fontdimen4\font\relax}
\providecommand{\BIBforeignlanguage}[2]{{%
\expandafter\ifx\csname l@#1\endcsname\relax
\typeout{** WARNING: IEEEtran.bst: No hyphenation pattern has been}%
\typeout{** loaded for the language `#1'. Using the pattern for}%
\typeout{** the default language instead.}%
\else
\language=\csname l@#1\endcsname
\fi
#2}}
\providecommand{\BIBdecl}{\relax}
\BIBdecl

\bibitem{Huang:2001:SLP:560905}
X.~Huang, A.~Acero, and H.-W. Hon, \emph{Spoken Language Processing: A Guide to
  Theory, Algorithm, and System Development}, 1st~ed.\hskip 1em plus 0.5em
  minus 0.4em\relax Upper Saddle River, NJ, USA: Prentice Hall PTR, 2001.

\bibitem{xue2014singular}
J.~Xue, J.~Li, D.~Yu, M.~Seltzer, and Y.~Gong, ``Singular value decomposition
  based low-footprint speaker adaptation and personalization for deep neural
  network,'' in \emph{Acoustics, Speech and Signal Processing (ICASSP), 2014
  IEEE International Conference on}.\hskip 1em plus 0.5em minus 0.4em\relax
  IEEE, 2014, pp. 6359--6363.

\bibitem{peddinti2018low}
V.~Peddinti, Y.~Wang, D.~Povey, and S.~Khudanpur, ``Low latency acoustic
  modeling using temporal convolution and lstms,'' \emph{IEEE Signal Processing
  Letters}, vol.~25, no.~3, pp. 373--377, 2018.

\bibitem{mcgraw2016personalized}
I.~McGraw, R.~Prabhavalkar, R.~Alvarez, M.~G. Arenas, K.~Rao, D.~Rybach,
  O.~Alsharif, H.~Sak, A.~Gruenstein, F.~Beaufays \emph{et~al.}, ``Personalized
  speech recognition on mobile devices,'' in \emph{Acoustics, Speech and Signal
  Processing (ICASSP), 2016 IEEE International Conference on}.\hskip 1em plus
  0.5em minus 0.4em\relax IEEE, 2016, pp. 5955--5959.

\bibitem{xiang2017binary}
X.~Xiang, Y.~Qian, and K.~Yu, ``Binary deep neural networks for speech
  recognition,'' \emph{Proc. Interspeech 2017}, pp. 533--537, 2017.

\bibitem{pundak2016lower}
G.~Pundak and T.~N. Sainath, ``Lower frame rate neural network acoustic
  models.'' in \emph{Interspeech}, 2016, pp. 22--26.

\bibitem{mohri2002weighted}
M.~Mohri, F.~Pereira, and M.~Riley, ``Weighted finite-state transducers in
  speech recognition,'' \emph{Computer Speech \& Language}, vol.~16, no.~1, pp.
  69--88, 2002.

\bibitem{hori2004fast}
T.~Hori, C.~Hori, and Y.~Minami, ``Fast on-the-fly composition for weighted
  finite-state transducers in 1.8 million-word vocabulary continuous speech
  recognition,'' in \emph{Eighth International Conference on Spoken Language
  Processing}, 2004.

\bibitem{soltau2009dynamic}
H.~Soltau and G.~Saon, ``Dynamic network decoding revisited,'' in
  \emph{Automatic Speech Recognition \& Understanding, 2009. ASRU 2009. IEEE
  Workshop on}.\hskip 1em plus 0.5em minus 0.4em\relax IEEE, 2009, pp.
  276--281.

\bibitem{nolden2012search}
D.~Nolden, R.~Schl{\"u}ter, and H.~Ney, ``Search space pruning based on
  anticipated path recombination in lvcsr,'' in \emph{Interspeech}, 2012.

\bibitem{sak2014long}
H.~Sak, A.~Senior, and F.~Beaufays, ``Long short-term memory recurrent neural
  network architectures for large scale acoustic modeling,'' in \emph{Fifteenth
  Annual Conference of the International Speech Communication Association},
  2014.

\bibitem{qian2016very}
Y.~Qian, M.~Bi, T.~Tan, and K.~Yu, ``Very deep convolutional neural networks
  for noise robust speech recognition,'' \emph{IEEE/ACM Transactions on Audio,
  Speech, and Language Processing}, vol.~24, no.~12, pp. 2263--2276, 2016.

\bibitem{chen2018progressive}
Z.~Chen, J.~Droppo, J.~Li, and W.~Xiong, ``Progressive joint modeling in
  unsupervised single-channel overlapped speech recognition,'' \emph{IEEE/ACM
  Transactions on Audio, Speech and Language Processing (TASLP)}, vol.~26,
  no.~1, pp. 184--196, 2018.

\bibitem{chen2018sequence}
Z.~Chen and J.~Droppo, ``Sequence modeling in unsupervised single-channel
  overlapped speech recognition.''\hskip 1em plus 0.5em minus 0.4em\relax
  ICASSP, 2018.

\bibitem{huang2018ctc}
M.~Huang, Y.~You, Z.~Chen, Y.~Qian, and K.~Yu, ``Knowledge distillation for
  sequence model.'' in \emph{Interspeech 2018}.

\bibitem{amodei2015deep}
D.~Amodei \emph{et~al.}, ``Deep speech 2: End-to-end speech recognition in
  english and mandarin,'' \emph{arXiv preprint arXiv:1512.02595}, 2015.

\bibitem{soltau2016neural}
H.~Soltau, H.~Liao, and H.~Sak, ``Neural speech recognizer: Acoustic-to-word
  lstm model for large vocabulary speech recognition,'' \emph{arXiv preprint
  arXiv:1610.09975}, 2016.

\bibitem{collobert2016wav2letter}
R.~Collobert, C.~Puhrsch, and G.~Synnaeve, ``Wav2letter: an end-to-end
  convnet-based speech recognition system,'' \emph{arXiv preprint
  arXiv:1609.03193}, 2016.

\bibitem{sak2015fast}
H.~Sak, A.~Senior, K.~Rao, and F.~Beaufays, ``Fast and accurate recurrent
  neural network acoustic models for speech recognition,'' \emph{arXiv preprint
  arXiv:1507.06947}, 2015.

\bibitem{chan2016end}
W.~Chan, ``End-to-end speech recognition models,'' Ph.D. dissertation, Carnegie
  Mellon University Pittsburgh, PA, 2016.

\bibitem{forney1973viterbi}
G.~D. Forney, ``The viterbi algorithm,'' \emph{Proceedings of the IEEE},
  vol.~61, no.~3, pp. 268--278, 1973.

\bibitem{povey2012generating}
D.~Povey, M.~Hannemann, G.~Boulianne, L.~Burget, A.~Ghoshal, M.~Janda,
  M.~Karafi{\'a}t, S.~Kombrink, P.~Motl{\'\i}{\v{c}}ek, Y.~Qian \emph{et~al.},
  ``Generating exact lattices in the wfst framework,'' in \emph{Acoustics,
  Speech and Signal Processing (ICASSP), 2012 IEEE International Conference
  on}.\hskip 1em plus 0.5em minus 0.4em\relax IEEE, 2012, pp. 4213--4216.

\bibitem{zhc00-chen-is16}
Z.~Chen, W.~Deng, T.~Xu, and K.~Yu, ``Phone synchronous decoding with ctc
  lattice,'' in \emph{Interspeech 2016}, 2016, pp. 1923--1927.

\bibitem{zhc00-chen-tasl2017}
Z.~Chen, Y.~Zhuang, Y.~Qian, and K.~Yu, ``Phone synchronous speech recognition
  with ctc lattices,'' \emph{IEEE/ACM Transactions on Audio, Speech, and
  Language Processing}, vol.~25, no.~1, pp. 86--97, Jan 2017.

\bibitem{graves2006connectionist}
A.~Graves, S.~Fern{\'a}ndez, F.~Gomez, and J.~Schmidhuber, ``Connectionist
  temporal classification: labelling unsegmented sequence data with recurrent
  neural networks,'' in \emph{Proceedings of the 23rd international conference
  on Machine learning}.\hskip 1em plus 0.5em minus 0.4em\relax ACM, 2006, pp.
  369--376.

\bibitem{povey2016purely}
D.~Povey, V.~Peddinti, D.~Galvez, P.~Ghahrmani, V.~Manohar, X.~Na, Y.~Wang, and
  S.~Khudanpur, ``Purely sequence-trained neural networks for asr based on
  lattice-free mmi,'' \emph{Submitted to Interspeech}, 2016.

\bibitem{godfrey1992switchboard}
J.~J. Godfrey, E.~C. Holliman, and J.~McDaniel, ``Switchboard: Telephone speech
  corpus for research and development,'' in \emph{Acoustics, Speech, and Signal
  Processing, 1992. ICASSP-92., 1992 IEEE International Conference on},
  vol.~1.\hskip 1em plus 0.5em minus 0.4em\relax IEEE, 1992, pp. 517--520.

\bibitem{chen2018gpu}
Z.~Chen, J.~Luitjens, H.~Xu, Y.~Wang, D.~Povey, and S.~Khudanpur, ``A gpu-based
  wfst decoder with exact lattice generation,'' \emph{arXiv preprint
  arXiv:1804.03243}, 2018.

\bibitem{povey2011kaldi}
D.~Povey, A.~Ghoshal, G.~Boulianne, L.~Burget, O.~Glembek, N.~Goel,
  M.~Hannemann, P.~Motlicek, Y.~Qian, P.~Schwarz \emph{et~al.}, ``The kaldi
  speech recognition toolkit,'' in \emph{IEEE 2011 workshop on automatic speech
  recognition and understanding}, no. EPFL-CONF-192584.\hskip 1em plus 0.5em
  minus 0.4em\relax IEEE Signal Processing Society, 2011.

\bibitem{woodland19951994}
S.~Young, ``The htk book version 3.4. 1,'' 2009.

\bibitem{cuda9}
``Cuda toolkit documentation,'' \url{http://docs.nvidia.com/cuda/}, accessed:
  2018-03-17.

\bibitem{e2e-2018}
Z.~Chen, Q.~Liu, H.~Li, and K.~Yu, ``On modular training of neural
  acoustics-to-word model for lvcsr,'' in \emph{ICASSP}, April 2018.

\bibitem{chen2017unified}
Z.~Chen, Y.~Qian, and K.~Yu, ``A unified confidence measure framework using
  auxiliary normalization graph,'' in \emph{International Conference on
  Intelligent Science and Big Data Engineering}.\hskip 1em plus 0.5em minus
  0.4em\relax Springer, 2017, pp. 123--133.

\bibitem{chen2017confidence}
Z.~Chen, Y.~Zhuang, and K.~Yu, ``Confidence measures for ctc-based phone
  synchronous decoding,'' in \emph{Acoustics, Speech and Signal Processing
  (ICASSP), 2017 IEEE International Conference on}.\hskip 1em plus 0.5em minus
  0.4em\relax IEEE, 2017, pp. 4850--4854.

\bibitem{chen2018kws}
Z.~Chen, Y.~Qian, and K.~Yu, ``Sequence discriminative training for deep
  learning based acoustic keyword spotting,'' \emph{Speech Communication},
  2018.

\bibitem{xu2017pruned}
H.~Xu, T.~Chen, D.~Gao, Y.~Wang, K.~Li, N.~Goel, Y.~Carmiel, D.~Povey, and
  S.~Khudanpur, ``A pruned rnnlm lattice-rescoring algorithm for automatic
  speech recognition,'' 2017.

\end{thebibliography}


\begin{thebibliography}{10}
\providecommand{\url}[1]{#1}
\csname url@samestyle\endcsname
\providecommand{\newblock}{\relax}
\providecommand{\bibinfo}[2]{#2}
\providecommand{\BIBentrySTDinterwordspacing}{\spaceskip=0pt\relax}
\providecommand{\BIBentryALTinterwordstretchfactor}{4}
\providecommand{\BIBentryALTinterwordspacing}{\spaceskip=\fontdimen2\font plus
\BIBentryALTinterwordstretchfactor\fontdimen3\font minus
  \fontdimen4\font\relax}
\providecommand{\BIBforeignlanguage}[2]{{%
\expandafter\ifx\csname l@#1\endcsname\relax
\typeout{** WARNING: IEEEtran.bst: No hyphenation pattern has been}%
\typeout{** loaded for the language `#1'. Using the pattern for}%
\typeout{** the default language instead.}%
\else
\language=\csname l@#1\endcsname
\fi
#2}}
\providecommand{\BIBdecl}{\relax}
\BIBdecl

\bibitem{xue2014singular}
J.~Xue, J.~Li, D.~Yu, M.~Seltzer, and Y.~Gong, ``Singular value decomposition
  based low-footprint speaker adaptation and personalization for deep neural
  network,'' in \emph{Acoustics, Speech and Signal Processing (ICASSP), 2014
  IEEE International Conference on}.\hskip 1em plus 0.5em minus 0.4em\relax
  IEEE, 2014, pp. 6359--6363.

\bibitem{peddinti2018low}
V.~Peddinti, Y.~Wang, D.~Povey, and S.~Khudanpur, ``Low latency acoustic
  modeling using temporal convolution and lstms,'' \emph{IEEE Signal Processing
  Letters}, vol.~25, no.~3, pp. 373--377, 2018.

\bibitem{mcgraw2016personalized}
I.~McGraw, R.~Prabhavalkar, R.~Alvarez, M.~G. Arenas, K.~Rao, D.~Rybach,
  O.~Alsharif, H.~Sak, A.~Gruenstein, F.~Beaufays \emph{et~al.}, ``Personalized
  speech recognition on mobile devices,'' in \emph{Acoustics, Speech and Signal
  Processing (ICASSP), 2016 IEEE International Conference on}.\hskip 1em plus
  0.5em minus 0.4em\relax IEEE, 2016, pp. 5955--5959.

\bibitem{xiang2017binary}
X.~Xiang, Y.~Qian, and K.~Yu, ``Binary deep neural networks for speech
  recognition,'' \emph{Proc. Interspeech 2017}, pp. 533--537, 2017.

\bibitem{pundak2016lower}
G.~Pundak and T.~N. Sainath, ``Lower frame rate neural network acoustic
  models.'' in \emph{Interspeech}, 2016, pp. 22--26.

\bibitem{zhc00-chen-tasl2017}
Z.~Chen, Y.~Zhuang, Y.~Qian, and K.~Yu, ``Phone synchronous speech recognition
  with ctc lattices,'' \emph{IEEE/ACM Transactions on Audio, Speech, and
  Language Processing}, vol.~25, no.~1, pp. 86--97, Jan 2017.

\bibitem{audhkhasi2017direct}
K.~Audhkhasi, B.~Ramabhadran, G.~Saon, M.~Picheny, and D.~Nahamoo, ``Direct
  acoustics-to-word models for english conversational speech recognition,''
  \emph{arXiv preprint arXiv:1703.07754}, 2017.

\bibitem{e2e-2018}
Z.~Chen, Q.~Liu, H.~Li, and K.~Yu, ``On modular training of neural
  acoustics-to-word model for lvcsr,'' in \emph{ICASSP}, April 2018.

\bibitem{mohri2002weighted}
M.~Mohri, F.~Pereira, and M.~Riley, ``Weighted finite-state transducers in
  speech recognition,'' \emph{Computer Speech \& Language}, vol.~16, no.~1, pp.
  69--88, 2002.

\bibitem{hori2004fast}
T.~Hori, C.~Hori, and Y.~Minami, ``Fast on-the-fly composition for weighted
  finite-state transducers in 1.8 million-word vocabulary continuous speech
  recognition,'' in \emph{Eighth International Conference on Spoken Language
  Processing}, 2004.

\bibitem{soltau2009dynamic}
H.~Soltau and G.~Saon, ``Dynamic network decoding revisited,'' in
  \emph{Automatic Speech Recognition \& Understanding, 2009. ASRU 2009. IEEE
  Workshop on}.\hskip 1em plus 0.5em minus 0.4em\relax IEEE, 2009, pp.
  276--281.

\bibitem{nolden2012search}
D.~Nolden, R.~Schl{\"u}ter, and H.~Ney, ``Search space pruning based on
  anticipated path recombination in lvcsr,'' in \emph{Thirteenth Annual
  Conference of the International Speech Communication Association}, 2012.

\bibitem{dixon2009harnessing}
P.~R. Dixon, T.~Oonishi, and S.~Furui, ``Harnessing graphics processors for the
  fast computation of acoustic likelihoods in speech recognition,''
  \emph{Computer Speech \& Language}, vol.~23, no.~4, pp. 510--526, 2009.

\bibitem{vesely2010parallel}
K.~Vesel{\`y}, L.~Burget, and F.~Gr{\'e}zl, ``Parallel training of neural
  networks for speech recognition,'' in \emph{International Conference on Text,
  Speech and Dialogue}.\hskip 1em plus 0.5em minus 0.4em\relax Springer, 2010,
  pp. 439--446.

\bibitem{you2009parallel}
K.~You, J.~Chong, Y.~Yi, E.~Gonina, C.~J. Hughes, Y.-K. Chen, W.~Sung, and
  K.~Keutzer, ``Parallel scalability in speech recognition,'' \emph{IEEE Signal
  Processing Magazine}, vol.~26, no.~6, 2009.

\bibitem{povey2011kaldi}
D.~Povey, A.~Ghoshal, G.~Boulianne, L.~Burget, O.~Glembek, N.~Goel,
  M.~Hannemann, P.~Motlicek, Y.~Qian, P.~Schwarz \emph{et~al.}, ``The kaldi
  speech recognition toolkit,'' in \emph{IEEE 2011 workshop on automatic speech
  recognition and understanding}, no. EPFL-CONF-192584.\hskip 1em plus 0.5em
  minus 0.4em\relax IEEE Signal Processing Society, 2011.

\bibitem{povey2012generating}
D.~Povey, M.~Hannemann, G.~Boulianne, L.~Burget, A.~Ghoshal, M.~Janda,
  M.~Karafi{\'a}t, S.~Kombrink, P.~Motl{\'\i}{\v{c}}ek, Y.~Qian \emph{et~al.},
  ``Generating exact lattices in the wfst framework,'' in \emph{Acoustics,
  Speech and Signal Processing (ICASSP), 2012 IEEE International Conference
  on}.\hskip 1em plus 0.5em minus 0.4em\relax IEEE, 2012, pp. 4213--4216.

\bibitem{mendis2016parallelizing}
C.~Mendis, J.~Droppo, S.~Maleki, M.~Musuvathi, T.~Mytkowicz, and G.~Zweig,
  ``Parallelizing wfst speech decoders,'' in \emph{Acoustics, Speech and Signal
  Processing (ICASSP), 2016 IEEE International Conference on}.\hskip 1em plus
  0.5em minus 0.4em\relax IEEE, 2016, pp. 5325--5329.

\bibitem{chong2010exploring}
J.~Chong, E.~Gonina, K.~You, and K.~Keutzer, ``Exploring recognition network
  representations for efficient speech inference on highly parallel
  platforms,'' in \emph{Eleventh Annual Conference of the International Speech
  Communication Association}, 2010.

\bibitem{kim2011h}
J.~Kim, K.~You, and W.~Sung, ``H-and c-level wfst-based large vocabulary
  continuous speech recognition on graphics processing units,'' in
  \emph{Acoustics, Speech and Signal Processing (ICASSP), 2011 IEEE
  International Conference on}.\hskip 1em plus 0.5em minus 0.4em\relax IEEE,
  2011, pp. 1733--1736.

\bibitem{kim2012efficient}
J.~Kim, J.~Chong, and I.~Lane, ``Efficient on-the-fly hypothesis rescoring in a
  hybrid gpu/cpu-based large vocabulary continuous speech recognition engine,''
  in \emph{Thirteenth Annual Conference of the International Speech
  Communication Association}, 2012.

\bibitem{kim2014accelerating}
J.~Kim and I.~Lane, ``Accelerating large vocabulary continuous speech
  recognition on heterogeneous cpu-gpu platforms,'' in \emph{Acoustics, Speech
  and Signal Processing (ICASSP), 2014 IEEE International Conference on}.\hskip
  1em plus 0.5em minus 0.4em\relax IEEE, 2014, pp. 3291--3295.

\bibitem{sak2010fly}
H.~Sak, M.~Saraclar, and T.~G{\"u}ng{\"o}r, ``On-the-fly lattice rescoring for
  real-time automatic speech recognition,'' in \emph{Eleventh Annual Conference
  of the International Speech Communication Association}, 2010.

\bibitem{argueta2017decoding}
A.~Argueta and D.~Chiang, ``Decoding with finite-state transducers on gpus,''
  \emph{arXiv preprint arXiv:1701.03038}, 2017.

\end{thebibliography}


\begin{thebibliography}{10}
\providecommand{\url}[1]{#1}
\csname url@samestyle\endcsname
\providecommand{\newblock}{\relax}
\providecommand{\bibinfo}[2]{#2}
\providecommand{\BIBentrySTDinterwordspacing}{\spaceskip=0pt\relax}
\providecommand{\BIBentryALTinterwordstretchfactor}{4}
\providecommand{\BIBentryALTinterwordspacing}{\spaceskip=\fontdimen2\font plus
\BIBentryALTinterwordstretchfactor\fontdimen3\font minus
  \fontdimen4\font\relax}
\providecommand{\BIBforeignlanguage}[2]{{%
\expandafter\ifx\csname l@#1\endcsname\relax
\typeout{** WARNING: IEEEtran.bst: No hyphenation pattern has been}%
\typeout{** loaded for the language `#1'. Using the pattern for}%
\typeout{** the default language instead.}%
\else
\language=\csname l@#1\endcsname
\fi
#2}}
\providecommand{\BIBdecl}{\relax}
\BIBdecl

\bibitem{xue2014singular}
J.~Xue, J.~Li, D.~Yu, M.~Seltzer, and Y.~Gong, ``Singular value decomposition
  based low-footprint speaker adaptation and personalization for deep neural
  network,'' in \emph{Acoustics, Speech and Signal Processing (ICASSP), 2014
  IEEE International Conference on}.\hskip 1em plus 0.5em minus 0.4em\relax
  IEEE, 2014, pp. 6359--6363.

\bibitem{mcgraw2016personalized}
I.~McGraw, R.~Prabhavalkar, R.~Alvarez, M.~G. Arenas, K.~Rao, D.~Rybach,
  O.~Alsharif, H.~Sak, A.~Gruenstein, F.~Beaufays \emph{et~al.}, ``Personalized
  speech recognition on mobile devices,'' in \emph{Acoustics, Speech and Signal
  Processing (ICASSP), 2016 IEEE International Conference on}.\hskip 1em plus
  0.5em minus 0.4em\relax IEEE, 2016, pp. 5955--5959.

\bibitem{pundak2016lower}
G.~Pundak and T.~N. Sainath, ``Lower frame rate neural network acoustic
  models.'' in \emph{Interspeech}, 2016, pp. 22--26.

\bibitem{zhc00-chen-is16}
Z.~Chen, W.~Deng, T.~Xu, and K.~Yu, ``Phone synchronous decoding with ctc
  lattice,'' in \emph{Interspeech 2016}, 2016, pp. 1923--1927.

\bibitem{zhc00-chen-tasl2017}
Z.~Chen, Y.~Zhuang, Y.~Qian, and K.~Yu, ``Phone synchronous speech recognition
  with ctc lattices,'' \emph{IEEE/ACM Transactions on Audio, Speech, and
  Language Processing}, vol.~25, no.~1, pp. 86--97, Jan 2017.

\bibitem{audhkhasi2017direct}
K.~Audhkhasi, B.~Ramabhadran, G.~Saon, M.~Picheny, and D.~Nahamoo, ``Direct
  acoustics-to-word models for english conversational speech recognition,''
  \emph{arXiv preprint arXiv:1703.07754}, 2017.

\bibitem{e2e-2018}
Z.~Chen, Q.~Liu, H.~Li, and K.~Yu, ``On modular training of neural
  acoustics-to-word model for lvcsr,'' in \emph{ICASSP}, April 2018.

\bibitem{mohri2002weighted}
M.~Mohri, F.~Pereira, and M.~Riley, ``Weighted finite-state transducers in
  speech recognition,'' \emph{Computer Speech \& Language}, vol.~16, no.~1, pp.
  69--88, 2002.

\bibitem{hori2004fast}
T.~Hori, C.~Hori, and Y.~Minami, ``Fast on-the-fly composition for weighted
  finite-state transducers in 1.8 million-word vocabulary continuous speech
  recognition,'' in \emph{Eighth International Conference on Spoken Language
  Processing}, 2004.

\bibitem{soltau2009dynamic}
H.~Soltau and G.~Saon, ``Dynamic network decoding revisited,'' in
  \emph{Automatic Speech Recognition \& Understanding, 2009. ASRU 2009. IEEE
  Workshop on}.\hskip 1em plus 0.5em minus 0.4em\relax IEEE, 2009, pp.
  276--281.

\bibitem{nolden2012search}
D.~Nolden, R.~Schl{\"u}ter, and H.~Ney, ``Search space pruning based on
  anticipated path recombination in lvcsr,'' in \emph{Interspeech}, 2012.

\bibitem{dixon2009harnessing}
P.~R. Dixon, T.~Oonishi, and S.~Furui, ``Harnessing graphics processors for the
  fast computation of acoustic likelihoods in speech recognition,''
  \emph{Computer Speech \& Language}, vol.~23, no.~4, pp. 510--526, 2009.

\bibitem{vesely2010parallel}
K.~Vesel{\`y}, L.~Burget, and F.~Gr{\'e}zl, ``Parallel training of neural
  networks for speech recognition,'' in \emph{International Conference on Text,
  Speech and Dialogue}.\hskip 1em plus 0.5em minus 0.4em\relax Springer, 2010,
  pp. 439--446.

\bibitem{you2009parallel}
K.~You, J.~Chong, Y.~Yi, E.~Gonina, C.~J. Hughes, Y.-K. Chen, W.~Sung, and
  K.~Keutzer, ``Parallel scalability in speech recognition,'' \emph{IEEE Signal
  Processing Magazine}, vol.~26, no.~6, 2009.

\bibitem{povey2011kaldi}
D.~Povey, A.~Ghoshal, G.~Boulianne, L.~Burget, O.~Glembek, N.~Goel,
  M.~Hannemann, P.~Motlicek, Y.~Qian, P.~Schwarz \emph{et~al.}, ``The kaldi
  speech recognition toolkit,'' in \emph{IEEE 2011 workshop on automatic speech
  recognition and understanding}, no. EPFL-CONF-192584.\hskip 1em plus 0.5em
  minus 0.4em\relax IEEE Signal Processing Society, 2011.

\bibitem{peddinti2018low}
V.~Peddinti, Y.~Wang, D.~Povey, and S.~Khudanpur, ``Low latency acoustic
  modeling using temporal convolution and lstms,'' \emph{IEEE Signal Processing
  Letters}, vol.~25, no.~3, pp. 373--377, 2018.

\bibitem{povey2012generating}
D.~Povey, M.~Hannemann, G.~Boulianne, L.~Burget, A.~Ghoshal, M.~Janda,
  M.~Karafi{\'a}t, S.~Kombrink, P.~Motl{\'\i}{\v{c}}ek, Y.~Qian \emph{et~al.},
  ``Generating exact lattices in the wfst framework,'' in \emph{Acoustics,
  Speech and Signal Processing (ICASSP), 2012 IEEE International Conference
  on}.\hskip 1em plus 0.5em minus 0.4em\relax IEEE, 2012, pp. 4213--4216.

\bibitem{mendis2016parallelizing}
C.~Mendis, J.~Droppo, S.~Maleki, M.~Musuvathi, T.~Mytkowicz, and G.~Zweig,
  ``Parallelizing wfst speech decoders,'' in \emph{Acoustics, Speech and Signal
  Processing (ICASSP), 2016 IEEE International Conference on}.\hskip 1em plus
  0.5em minus 0.4em\relax IEEE, 2016, pp. 5325--5329.

\bibitem{chong2010exploring}
J.~Chong, E.~Gonina, K.~You, and K.~Keutzer, ``Exploring recognition network
  representations for efficient speech inference on highly parallel
  platforms,'' in \emph{Interspeech}, 2010.

\bibitem{kim2011h}
J.~Kim, K.~You, and W.~Sung, ``H-and c-level wfst-based large vocabulary
  continuous speech recognition on graphics processing units,'' in
  \emph{Acoustics, Speech and Signal Processing (ICASSP), 2011 IEEE
  International Conference on}.\hskip 1em plus 0.5em minus 0.4em\relax IEEE,
  2011, pp. 1733--1736.

\bibitem{kim2012efficient}
J.~Kim, J.~Chong, and I.~Lane, ``Efficient on-the-fly hypothesis rescoring in a
  hybrid gpu/cpu-based large vocabulary continuous speech recognition engine,''
  in \emph{Thirteenth Annual Conference of the International Speech
  Communication Association}, 2012.

\bibitem{kim2014accelerating}
J.~Kim and I.~Lane, ``Accelerating large vocabulary continuous speech
  recognition on heterogeneous cpu-gpu platforms,'' in \emph{Acoustics, Speech
  and Signal Processing (ICASSP), 2014 IEEE International Conference on}.\hskip
  1em plus 0.5em minus 0.4em\relax IEEE, 2014, pp. 3291--3295.

\bibitem{sak2010fly}
H.~Sak, M.~Saraclar, and T.~G{\"u}ng{\"o}r, ``On-the-fly lattice rescoring for
  real-time automatic speech recognition,'' in \emph{Eleventh Annual Conference
  of the International Speech Communication Association}, 2010.

\bibitem{argueta2017decoding}
A.~Argueta and D.~Chiang, ``Decoding with finite-state transducers on gpus,''
  \emph{arXiv preprint arXiv:1701.03038}, 2017.

\bibitem{woodland19951994}
S.~Young, ``The htk book version 3.4. 1,'' 2009.

\bibitem{cuda9}
``Cuda toolkit documentation,'' \url{http://docs.nvidia.com/cuda/}, accessed:
  2018-03-17.

\bibitem{lamport1979make}
L.~Lamport, ``How to make a multiprocessor computer that correctly executes
  multiprocess progranm,'' \emph{IEEE transactions on computers}, no.~9, pp.
  690--691, 1979.

\bibitem{alakeel2010}
A.~M. Alakeel, ``A guide to dynamic load balancing in distributed computer
  systems,'' in \emph{International Journal of Computer Science and Network
  Security (IJCSNS}.\hskip 1em plus 0.5em minus 0.4em\relax Citeseer, 2010.

\bibitem{mangu1999finding}
L.~Mangu, E.~Brill, and A.~Stolcke, ``Finding consensus among words:
  Lattice-based word error minimization,'' in \emph{Sixth European Conference
  on Speech Communication and Technology}, 1999.

\bibitem{goel2000minimum}
V.~Goel and W.~J. Byrne, ``Minimum bayes-risk automatic speech recognition,''
  \emph{Computer Speech \& Language}, vol.~14, no.~2, pp. 115--135, 2000.

\bibitem{fiscus1997post}
J.~G. Fiscus, ``A post-processing system to yield reduced word error rates:
  Recognizer output voting error reduction (rover),'' in \emph{Automatic Speech
  Recognition and Understanding, 1997. Proceedings., 1997 IEEE Workshop
  on}.\hskip 1em plus 0.5em minus 0.4em\relax IEEE, 1997, pp. 347--354.

\bibitem{povey2005discriminative}
D.~Povey, ``Discriminative training for large vocabulary speech recognition,''
  Ph.D. dissertation, University of Cambridge, 2005.

\bibitem{ljolje1999efficient}
A.~Ljolje, F.~Pereira, and M.~Riley, ``Efficient general lattice generation and
  rescoring,'' in \emph{Sixth European Conference on Speech Communication and
  Technology}, 1999.

\bibitem{povey2016purely}
D.~Povey, V.~Peddinti, D.~Galvez, P.~Ghahrmani, V.~Manohar, X.~Na, Y.~Wang, and
  S.~Khudanpur, ``Purely sequence-trained neural networks for asr based on
  lattice-free mmi,'' \emph{Submitted to Interspeech}, 2016.

\bibitem{sak2014long}
H.~Sak, A.~Senior, and F.~Beaufays, ``Long short-term memory recurrent neural
  network architectures for large scale acoustic modeling,'' in \emph{Fifteenth
  Annual Conference of the International Speech Communication Association},
  2014.

\bibitem{hoffmeister2006frame}
B.~Hoffmeister, T.~Klein, R.~Schl{\"u}ter, and H.~Ney, ``Frame based system
  combination and a comparison with weighted rover and cnc.'' in
  \emph{INTERSPEECH}.\hskip 1em plus 0.5em minus 0.4em\relax Citeseer, 2006.

\bibitem{siu1999evaluation}
M.~Siu and H.~Gish, ``Evaluation of word confidence for speech recognition
  systems,'' \emph{Computer Speech \& Language}, vol.~13, no.~4, pp. 299--319,
  1999.

\bibitem{chen2017confidence}
Z.~Chen, Y.~Zhuang, and K.~Yu, ``Confidence measures for ctc-based phone
  synchronous decoding,'' in \emph{Acoustics, Speech and Signal Processing
  (ICASSP), 2017 IEEE International Conference on}.\hskip 1em plus 0.5em minus
  0.4em\relax IEEE, 2017, pp. 4850--4854.

\bibitem{chen2017unified}
Z.~Chen, Y.~Qian, and K.~Yu, ``A unified confidence measure framework using
  auxiliary normalization graph,'' in \emph{International Conference on
  Intelligent Science and Big Data Engineering}.\hskip 1em plus 0.5em minus
  0.4em\relax Springer, 2017, pp. 123--133.

\end{thebibliography}


\end{document}